\documentclass{article}

\usepackage[final]{nips18}
\usepackage[utf8]{inputenc} 
\usepackage[T1]{fontenc}    
\usepackage{hyperref}       
\usepackage{url}            
\usepackage{booktabs}       
\usepackage{amsmath}
\usepackage{nicefrac}       
\usepackage{microtype}      
\usepackage{authblk}
\usepackage{graphicx,accents}
\usepackage{color}
\usepackage[dvipsnames]{xcolor}
 \usepackage{float}
\usepackage{eurosym}

\DeclareMathOperator*{\argmin}{arg\,min}

\makeatletter
\DeclareRobustCommand{\cev}[1]{%
  \mathpalette\do@cev{#1}%
}
\newcommand{\do@cev}[2]{%
  \fix@cev{#1}{+}%
  \reflectbox{$\m@th#1\vec{\reflectbox{$\fix@cev{#1}{-}\m@th#1#2\fix@cev{#1}{+}$}}$}%
  \fix@cev{#1}{-}%
}
\newcommand{\fix@cev}[2]{%
  \ifx#1\displaystyle
    \mkern#23mu
  \else
    \ifx#1\textstyle
      \mkern#23mu
    \else
      \ifx#1\scriptstyle
        \mkern#22mu
      \else
        \mkern#22mu
      \fi
    \fi
  \fi
}

\usepackage{graphicx} 
\graphicspath{ {./images/} }

\title{Pruning and Sparsemax Methods \\ for Hierarchical Attention Networks}

\author[1,2]{João G. Ribeiro}
\author[1]{Frederico S. Felisberto}
\author[1]{Isabel C. Neto}

\affil[1]{Department of Computer Science, Instituto Superior Técnico, University of Lisbon}
\affil[2]{INESC-ID}

\begin{document}

\maketitle

\begin{abstract}
    This paper introduces and evaluates two novel Hierarchical Attention Network models \citep{yang2016hierarchical} - i) Hierarchical Pruned Attention Networks, which remove the irrelevant words and sentences from the classification process in order to reduce potential noise in the document classification accuracy and ii) Hierarchical Sparsemax Attention Networks, which replace the Softmax function used in the attention mechanism with the Sparsemax \citep{martins2016softmax}, capable of better handling importance distributions where a lot of words or sentences have very low probabilities. Our empirical evaluation on the IMDB Review for sentiment analysis datasets shows both approaches to be able to match the results obtained by the current state-of-the-art (without, however, any significant benefits). All our source code is made available at \url{https://github.com/jmribeiro/dsl-project}.
\end{abstract}

\section{Introduction}

Recent years have witnessed a significant number of successes in the sub-field of natural language processing, document classification \citep{yang2016hierarchical, kowsari2019text, kowsari2018rmdl, chen2018verbal, jiang2018text}. 

With the introduction of \emph{Hierarchical Attention Networks} (HANs) \citep{yang2016hierarchical}, the classification of documents while identifying which words and sentences are or not relevant for the task turned out to be a crucial issue in the field. These models don't only take into account a document's hierarchical structure (i.e., documents contain sentences and sentences contain words), but also introduce an \emph{attention mechanism} which assigns an relevance value to the words and sentences, telling how relevant they are for the task at hand. \cite{yang2016hierarchical} evaluate their approach on six different datasets (Yelp 2013, 2014 and 2015, IMDB review, Yahoo Answer and Amazon Review \citep{yang2016hierarchical}) and are able to obtain state-of-the-art document classification accuracies, outperforming all previous works.

However, such models always assume these components (words and sentences) to be at least somewhat relevant for classification (when some components may only be introducing noise into the discrimination process itself). In this paper we therefore follow \cite{yang2016hierarchical} by going one step further and asking a new hypothesis - what if words and sentences, which the attention mechanism assigns a very low value (e.g., stop words), instead of being assigned low values, could be instead completely removed from the classification process?

To tackle this issue, this paper introduces and evaluates two new methods. The first one, named the \emph{Hierarchical Pruned Attention Networks} aims to ignore words and sentences which may not be very relevant for the classification, and the second one, named the \emph{Hierarchical Sparsemax Attention Networks} replaces the Softmax function in the attention mechanism in order to better handle a lot of irrelevant words and sentences. We perform an empirical evaluation (N=5, confidence level of 99\%), comparing both models against the original Hierarchical Attention Networks. Our evaluation on the IMDB review dataset for sentiment analysis \citep{maas-EtAl:2011:ACL-HLT2011} shows possible to match the current state-of-the-art, however without any clear benefits of choosing one method over the others.

\section{Outline}

This rest of this paper is organized as following:

\begin{itemize}
    \item Section \ref{sec:related-work} surveys the relevant literature.
    \item Section \ref{sec:background} details the hierarchical attention networks architecture, which sets the foundation for our two novel models.
    \item Section \ref{sec:architecture} highlights our main contributions and describes the two main architectures introduced.
    \item Section \ref{sec:evaluation} describes the performed empirical evaluation procedure.
    \item Section \ref{sec:results} reports and discusses the obtained results.
    \item Section \ref{sec:conclusion} concludes the paper and lists future lines of work.
\end{itemize}

\section{Related Work}
\label{sec:related-work}

Available literature in the NLP sub-field of document classification can be surveyed in the light of the document classification accuracy \citep{kowsari2019text}.

When the field of natural language processing was benefiting from the usage of linear models combined with tools, such as bag-of-words and n-grams, \cite{zhang2015character} first introduce LSTMs and CNNs for text classification. These models, still having a rather low degree of complexity, are shown by the authors to outperform traditional models in three datasets (Yelp'15, Yahoo Answers and Amazon).

Following their line of work, \cite{tang2015document} later on introduce gated recurrent neural networks for document classification. Their models outperform the current state-of-the-art at the time, in the Yelp'15 dataset and are also evaluated on three additional datasets - the Yelp'13, Yelp'14 and IMDB datasets \citep{tang2015document}.

Finally, following \cite{tang2015document}, \citet{yang2016hierarchical} introduced Hierarchical Attention Networks (HANs).
Unlike most document classifiers, HANs assume a hierarchical structure to a document (i.e., a document is a sequence of sentences, and sentences are sequences of words). Instead of processing a document as a simple sequence of words, HANs exploit this hierarchical structure while weighting the relevance of each word/sentence using an attention mechanism (hence the name).

\cite{yang2016hierarchical} then evaluate their model on six different datasets (Yelp 2013, 2014 and 2015), IMDB, Yahoo Answer, and Amazon \citep{yang2016hierarchical}) and are able to obtain state-of-the-art document classification accuracies, outperforming all previous works.

\subsection*{Discussion}

Taking all the relevant literature into account, we follow \citet{yang2016hierarchical} and ask the following question - what if classification accuracy could be improved by either removing all words/sentences with low attention weights or using a function other than the original Softmax? Recently, \cite{martins2016softmax} introduced the Sparsemax function which is capable of better handling distributions with a lot of near-zero probabilities.
When there are a lot of words/sentences which may not be relevant to the classification task at all (such as stop words, assuming such components to be always somewhat relevant for the classification task may introduce noise into the process. We therefore extend the available literature by introducing and evaluating two novel models which tackle this issue. 

\clearpage

\section{Hierarchical Attention Networks (Background)}
\label{sec:background}

\begin{figure}[H]
    \centering
    \includegraphics[width=0.6\linewidth]{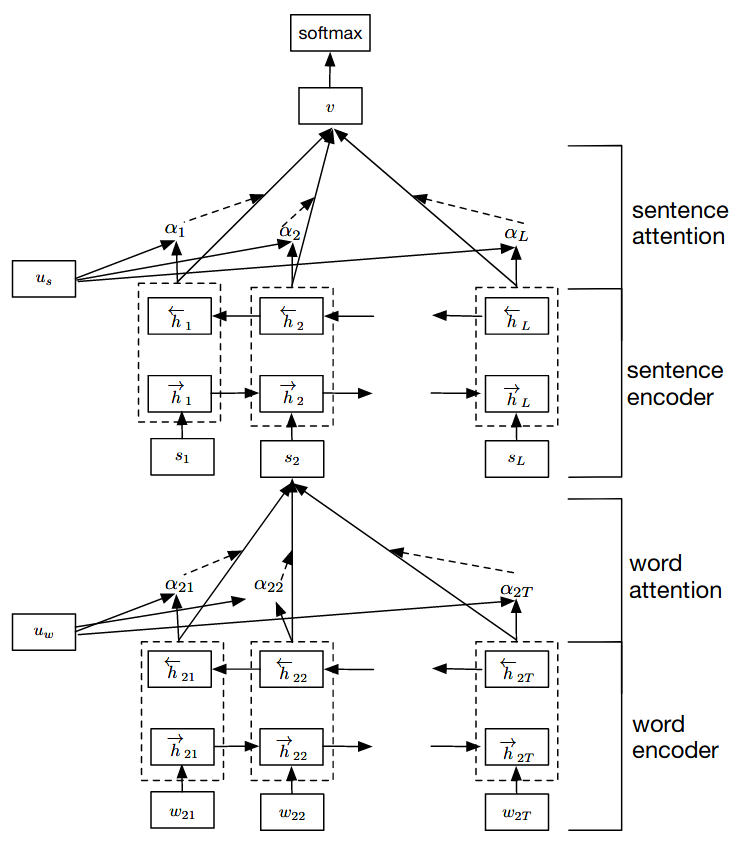}
    \caption{The Hierarchical Attention Network architecture (from \citep{yang2016hierarchical}).}
    \label{fig:han}
\end{figure}

Hierarchical Attention Networks (HANs) are models which discriminate a document into its most likely class.
An overview of their architecture can be visualized in figure \ref{fig:han}.

Given as input a flat, padded document $d$ describing a sequence of word tokens/indices from a given vocabulary, with shape $(S, L)$ (where $S$ is the number of sentences and $L$ the number of words per sentence), the classification process is detailed as following:

\begin{enumerate}
    
    \item Each sentence of $L$ words is first passed through a embedding layer, obtaining a word embeddings tensor $w$ with shape $(S, L, E)$, where $E$ represents the dimension of a single embedding vector. As in the original work, we resort to a pre-loaded weight vector with dimension of 200 embeddings, however, unlike the original work, we do not pre-train this vectors ourselves (loading a pre-trained GloVe \footnote{https://nlp.stanford.edu/projects/glove/} vector)
    
    \item Word embeddings $w$ are then passed through a bidirectional GRU, computing the hidden word scores $h = [\overrightarrow{h}, \overleftarrow{h}]$.
    
    \item Hidden word scores $h$ are then passed through a single-layer feedforward network, obtaining hidden word representations $u = tanh(W_w h + b_w)$.
    
    \item Hidden word representations are then passed through an attention mechanism, which using a trainable word level context vector $u_w$, compute the attention weights $\alpha = Softmax(u^\top u_w)$ 
    
    \item Attention weights $\alpha$ and hidden word scores $h$ are multiplied, obtaining the sentence vectors $s = \alpha \odot h$
    
    \item Sentence vectors $s$ are then passed through a bidirectional GRU, computing the hidden sentence scores $h_s = [\overrightarrow{h_s}, \overleftarrow{h_s}]$.
    
    \item Hidden sentence scores $h_s$ are then passed through a single-layer feedforward network, obtaining hidden sentence representations $u_s = tanh(W_s h_s + b_s)$.
    
    \item Hidden sentence representations are then passed through an attention mechanism, which using a trainable sentence level context vector $u_s$, compute the attention weights $\alpha = Softmax(u_s^\top u_s)$ 
    
    \item Attention weights $\alpha$ and hidden sentence scores $h_s$ are then multiplied, obtaining the document features $v = \alpha \odot h_s$
    
    \item Document features $v$ are finally passed through an affite transformation, obtaining the class logit scores $z = W_v v + b_v$
    
    \item Finally, $z$ can be used to compute the probabilities of each class $p=softmax(z)$. Cross entropy loss is then used to compute the gradients for all parameters.
    
\end{enumerate}

\section{Proposed Models}
\label{sec:architecture}

We introduce two novel model architectures to tackle the issue of unimportant words and sentences in the original attention mechanisms \citet{yang2016hierarchical}. The first one, which we named \emph{Hierarchical Pruned Attention Networks} (HPAN), completely ignores components with low attention weights, and the second, named \emph{Hierarchical Sparsemax Attention Networks} (HSAN), replaces the original Softmax function with the Sparsemax function, in order to better handle cases when there are a lot of irrelevant components (in which case simply removing them using the first approach may not be helpful).

\subsection{Hierarchical Pruned Attention Networks (HPAN)}

However, given that all components are taken into account, components with very low attention values (such as stop words), may only be introducing noise into the classification process. To mitigate this issue we modify both attention mechanisms to completely ignore words and sentences with an attention weight below a given threshold (which we call $\alpha_{min}$). 

Like with the original HANs, the HPANs, given as input a flat, padded document $d$ describing a sequence of word tokens/indices from a given vocabulary, with shape $(S, L)$ (where $S$ is the number of sentences and $L$ the number of words per sentence), follow the detailed classification process:

\begin{enumerate}
    
    \item Each sentence of $L$ words is first passed through a embedding layer, obtaining a word embeddings tensor $w$ with shape $(S, L, E)$, where $E$ represents the dimension of a single embedding vector. As in the original work, we resort to a pre-loaded weight vector with dimension of 200 embeddings (GloVe).
    
    \item Word embeddings $w$ are then passed through a bidirectional GRU, computing the hidden word scores $h = [\overrightarrow{h}, \overleftarrow{h}]$.
    
    \item Hidden word scores $h$ are then passed through a single-layer feedforward network, obtaining hidden word representations $u = tanh(W_w h + b_w)$.
    
    \item Hidden word representations are then passed through an attention mechanism, which using a trainable word level context vector $u_w$, compute the attention weights $\alpha = Softmax(u^\top u_w)$ 
    
    \item Attention weights $\alpha$ below a minimum attention threshold $\alpha_{min}$ are set to zero and then normalized so that the remaining attention weights still sum up to 1.0.
    
    \item New attention weights $\alpha$ and hidden word scores $h$ are then multiplied, obtaining the sentence vectors $s = \alpha \odot h$
    
    \item Sentence vectors $s$ are then passed through a bidirectional GRU, computing the hidden sentence scores $h_s = [\overrightarrow{h_s}, \overleftarrow{h_s}]$.
    
    \item Hidden sentence scores $h_s$ are then passed through a single-layer feedforward network, obtaining hidden sentence representations $u_s = tanh(W_s h_s + b_s)$.
    
    \item Hidden sentence representations are then passed through an attention mechanism, which using a trainable sentence level context vector $u_s$, compute the attention weights $\alpha = Softmax(u_s^\top u_s)$ 
    
    \item Attention weights $\alpha$ below a minimum attention threshold $\alpha_{min}$ are set to zero and then normalized so that the remaining attention weights still sum up to 1.0.
    
    \item New attention weights $\alpha$ and hidden sentence scores $h_s$ are then again multiplied, obtaining the document features $v = \alpha \odot h_s$
    
    \item Document features $v$ are finally passed through an affite transformation, obtaining the class logit scores $z = W_v v + b_v$
    
    \item Finally, $z$ can be used to compute the probabilities of each class $p=softmax(z)$. Cross entropy loss is then used to compute the gradients for all parameters.
    
\end{enumerate}

\subsection{Hierarchical Sparsemax Attention Networks (HSAN)}

In the original HANs \cite{yang2016hierarchical}, attention weights $\alpha$ are computed using the Softmax function:

\begin{equation} \label{eq:softmax}
    softmax(z) := \frac{exp(z)}{\sum_{z'}exp(z')}
\end{equation}

However, if there are a lot of irrelevant words and sentences, completely removing them using the pruning mechanism described above may not be beneficial. We therefore follow \cite{martins2016softmax} work by using their introduced Sparsemax function. 

The Sparsemax function is proven to better handle distributions with a lot of near-zero probabilities by returning the \textit{"Euclidean projection of the input vector z onto the probability simplex"} \citep{martins2016softmax}, i.e.:

\begin{equation} \label{eq:sparsemax}
    sparsemax(z) := \argmin_{\mathbf{p} \in \Delta^{k+1}} ||\mathbf{p}-z||^2
\end{equation}

The combination of such function with the attention mechanism only seemed logical, and therefore, to mitigate this issue, we explore a second model, named the Hierarchical Sparsemax Attention Network (HSAN), which replaces the Softmax function used in both attention mechanisms with the Sparsemax function.

Once again, given as input a flat, padded document $d$ describing a sequence of word tokens/indices from a given vocabulary, with shape $(S, L)$ (where $S$ is the number of sentences and $L$ the number of words per sentence), the document classification process for the HSANs is the following:

\begin{enumerate}
    
    \item Each sentence of $L$ words is first passed through a embedding layer, obtaining a word embeddings tensor $w$ with shape $(S, L, E)$, where $E$ represents the dimension of a single embedding vector. As in the original work, we resort to a pre-loaded weight vector with dimension of 200 embeddings (GloVe).
    
    \item Word embeddings $w$ are then passed through a bidirectional GRU, computing the hidden word scores $h = [\overrightarrow{h}, \overleftarrow{h}]$.
    
    \item Hidden word scores $h$ are then passed through a single-layer feedforward network, obtaining hidden word representations $u = tanh(W_w h + b_w)$.
    
    \item Hidden word representations are then passed through an attention mechanism, which using a trainable word level context vector $u_w$, compute the attention weights $\alpha = sparsemax(u^\top u_w)$ 
    
    \item Attention weights $\alpha$ and hidden word scores $h$ are multiplied, obtaining the sentence vectors $s = \alpha \odot h$
    
    \item Sentence vectors $s$ are then passed through a bidirectional GRU, computing the hidden sentence scores $h_s = [\overrightarrow{h_s}, \overleftarrow{h_s}]$.
    
    \item Hidden sentence scores $h_s$ are then passed through a single-layer feedforward network, obtaining hidden sentence representations $u_s = tanh(W_s h_s + b_s)$.
    
    \item Hidden sentence representations are then passed through an attention mechanism, which using a trainable sentence level context vector $u_s$, compute the attention weights $\alpha = sparsemax(u_s^\top u_s)$ 
    
    \item Attention weights $\alpha$ and hidden sentence scores $h_s$ are then multiplied, obtaining the document features $v = \alpha \odot h_s$
    
    \item Document features $v$ are finally passed through an affite transformation, obtaining the class logit scores $z = W_v v + b_v$
    
    \item Finally, $z$ can be used to compute the probabilities of each class $p=softmax(z)$. Cross entropy loss is then used to compute the gradients for all parameters.
    
\end{enumerate}

\section{Evaluation}
\label{sec:evaluation}

We compare both our models (HPAN and HSNA) against the original HAN.

Table \ref{tab:hyper} displays all used hyperparameters, alongside their descriptions and their values.

\begin{table}[H]
    \centering
    \caption{Hyperparameter values used when training each model.}
    \begin{tabular}{rrll}
    \toprule
    Hyperparameter & Our value & \cite{yang2016hierarchical}'s value \\
    \midrule
    Word2vec embeddings size & 200 & 200\\
    GRU layers & 1 & 1\\
    GRU layers hidden sizes & 50 & 50 \\
    Dropout & 0.1 & unreported \\
    Training epochs & 3 & unreported \\
    Optimizer & adam & sgd \\
    Learning rate & 0.001 & unreported* \\
    Batch size & 64 & 64 \\
    HPAN min. attention threshold & 0.05 & n.a. \\
    \bottomrule
    \end{tabular}
    \label{tab:hyper}
\end{table}

\noindent * The authors pick the best learning rate from a grid search, however, do not report the value. \citep{yang2016hierarchical}.

\subsection{Datasets}

Unfortunately, due to long loading and training times for most of the datasets, we decided to perform our evaluation for all models in the IMDB Review dataset for sentiment analysis \footnote{https://ai.stanford.edu/~amaas/data/sentiment/}.

Even though we still provide all the necessary code to train any model on any of the original datasets (Yelp'15, Yahoo Answers, Amazon and IMDB), we chose this dataset due to its relatively low sample size (which allows both faster loading and training) and its simplicity.

The IMDB review dataset, introduced by \cite{maas-EtAl:2011:ACL-HLT2011} contains 50'000 movie review and polarizes their classifications to either positive and negative (unlike the original IMDB dataset which kept 10 classification labels). There's an even number of positive and negative reviews. However, the authors \citep{maas-EtAl:2011:ACL-HLT2011} only considered \textit{'highly polarized reviews'}, or in other words, classifications are considered positive a movie review with a classification of 7 or higher, and negative a movie review with a classification of 4 or lower (both out of 10). Movie reviews with classifications of 5 and 6 were not included. The dataset was then evenly divided into training and test sets, both containing 25'000 movie reviews, and the 25'000 training reviews were then split into 70\% training and 30\% validation.

We load the IMDB review dataset and preprocess it using the official Pytorch torchtext module \footnote{https://pytorch.org/text/\_modules/torchtext/datasets/imdb.html}, using the spacy tokenizer \footnote{https://spacy.io/api/tokenizer}.

\subsection{Experimental Procedure}

Given a total of three models - i) HAN, ii) HPAN and iii) HSAN and a single dataset - IMDB review for sentiment analysis - the evaluation procedure for a single experiment for a model (N=1) was the following:

\begin{enumerate}
    \item Train a model on the IMDB dataset for three epochs on the training set.
    \item Evaluate, at each epoch, the classification accuracy on the validation set.
    \item Evaluate, after all epochs, the classification accuracy on test dataset.
\end{enumerate}

Finally, to compare the results obtained by our models, we use the following metric:

\textbf{M}: Document classification accuracy (\%) on obtained by a trained model on the test dataset, after three training epochs. 

\section{Results}
\label{sec:results}

All three models (HAN, HPAN and HSAN) were trained and evaluated according to metric \textbf{M} (classification accuracy (\%) on test dataset, after three training epochs).

Once again, due to time constraints and available computational resources, we were able to perform a total o 5 (independent) experiments (N=5) for the HAN and HSAN models and 3 experiments (N=3) for the HPAN model (due to gradient computation issue, further discussed in section \ref{sec:conclusion}).

\begin{figure}
    \centering
    \includegraphics[width=0.7\linewidth]{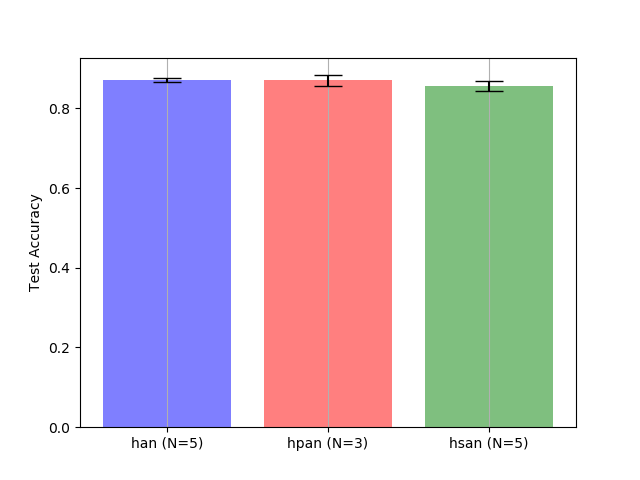}
    \caption{Mean test accuracies obtained after three training epochs over 5 total experiments. Errors computed using a confidence level of 99\%.}
    \label{fig:results-bar}
\end{figure}

Table \ref{tab:results} reports the obtained document classification accuracies on test sets for all models and figure \ref{fig:results-bar} reports our obtained results with a confidence level of 99\%.

\begin{table}[H]
    \centering
    \caption{Obtained document classification accuracies in the IMDB review test set, for all three models, after three training epochs.}
    \begin{tabular}{cccc}
    \toprule
    \textbf{Model} & $\mu$ & $\sigma$ & \textbf{N}\\
    \midrule
    HAN & 87.08\% & 0.004 & 5\\
    HPAN & 87.01\% & 0.011 & 3\\
    HSAN & 85.64\% & 0.011 & 5\\
    \bottomrule
    \end{tabular}
    \label{tab:results}
\end{table}

Taking all experiments into account, we can make three conclusions. 

First, both our models were able to perform similarly as the current state-of-the-art (HAN). The second conclusion is that neither pruning words and sentences nor replacing the Softmax function with the Sparsemax function in the attention mechanism showed great advantage. 

Both these conclusions are only true regarding the fact that the experiments were conducted on the IMDB Review dataset for sentiment analysis. This binary/polarized dataset, even tough it provided a first good test bed to evaluate our proposed models, may not allow deeper conclusions due to its simplicity. The only way to extrapolate these conclusions is to repeat the process by evaluating all three models using more complex datasets (such as the ones used in the original work \citep{yang2016hierarchical}).

Finally, the HSAN hinted a slightly worst performance after training for the same number of epochs and using the same hyperparameters as the other models. There wasn't, however, a very significant difference to make this conclusion a very strong one.

\section{Conclusion \& Future Work}
\label{sec:conclusion}

In this paper we tackled the issue of irrelevant words and sentences by modifying the Hierarchical Attention Networks attention mechanism \citep{yang2016hierarchical}. We introduced and evaluated two novel models, namely the Hierarchical Pruned Attention Networks and the Hierarchical Sparsemax Attention Networks. We compared both models against the original Hierarchical Attention Networks and, regarding the IMDB Review dataset for sentiment analysis, our empirical evaluation (N=5, confidence level of 99\%) showed there was no significant performance gains (in terms of document classification accuracy).

Unfortunately, the evaluation of these on a single, polarized dataset may not be enough to conclude that there are no advantages/disadvantages between the three methods. Logically, our next line of work will be to continue this study and perform a new, empirical evaluation, on three additional datasets -- Yelp'15, Amazon and Yahoo Answers -- which all label each document with one of K classes (instead of simply two, like the IMDB review for sentiment analysis). During this study, we will also set out to perform a fine-tuning for all hyperparameters (including the minimum attention threshold for the HPAN). 

Even though our HPAN model was able to train properly, it had the requirement of being restricted to a cpu device only, rendering its experiments very slow when in comparison with the other two models. This was caused by an issue in PyTorch's autograd module when running on a gpu device (and also happened sometimes when using cpu), which neither of us were able to solve - when setting to zero all attention weights below a minimum threshold, the backpropagation of the gradients yielded \emph{nan}.

Finally, it will also be interesting to study which are the effects, in terms of percentage of pruned words/sentences, when given a certain minimum attention threshold to the HPAN. Since there are two individual attention mechanisms (one for words and one for sentences), it can also prove interesting to explore the effects of pruning on each individual level (words/sentences). The same process can be done with the Sparsemax function, which is currently used on both levels.

\bibliographystyle{unsrtnat}
\bibliography{bibliography}

\end{document}